\documentclass[10pt,twocolumn,letterpaper]{article}

\usepackage[pagenumbers]{iccv} 

\usepackage{multirow}
\usepackage{pifont}
\usepackage{xcolor}
\usepackage{makecell}
\usepackage{colortbl}
\usepackage{arydshln}
\usepackage{cuted}
\usepackage{capt-of}
\usepackage{booktabs}
\usepackage{amssymb}
\usepackage[misc]{ifsym} 
\definecolor{iccvblue}{rgb}{0.21,0.49,0.74}
\usepackage[pagebackref,breaklinks,colorlinks,allcolors=iccvblue]{hyperref}
\def\confName{ICCV}

\newcommand{\methodname}{AnyI2V\xspace}

\title{AnyI2V: Animating Any Conditional Image with Motion Control}

\author{
Ziye Li$^1$
\quad
Hao Luo$^{2,3}$
\quad
Xincheng Shuai$^1$
\quad
Henghui Ding$^1$~\!$^{\textrm{\Letter}}$
\\
$^1$Fudan University
\quad
$^2$DAMO Academy, Alibaba group
\quad
$^3$Hupan Lab\\
\href{https://henghuiding.com/AnyI2V/}{https://henghuiding.com/AnyI2V/}
\vspace{-0.5cm}
}

\begin{document}
\twocolumn[{%
\renewcommand\twocolumn[1][]{#1}%
\maketitle
\begin{center}
    \centering
    \captionsetup{type=figure}
    \includegraphics[width=1.0\linewidth]{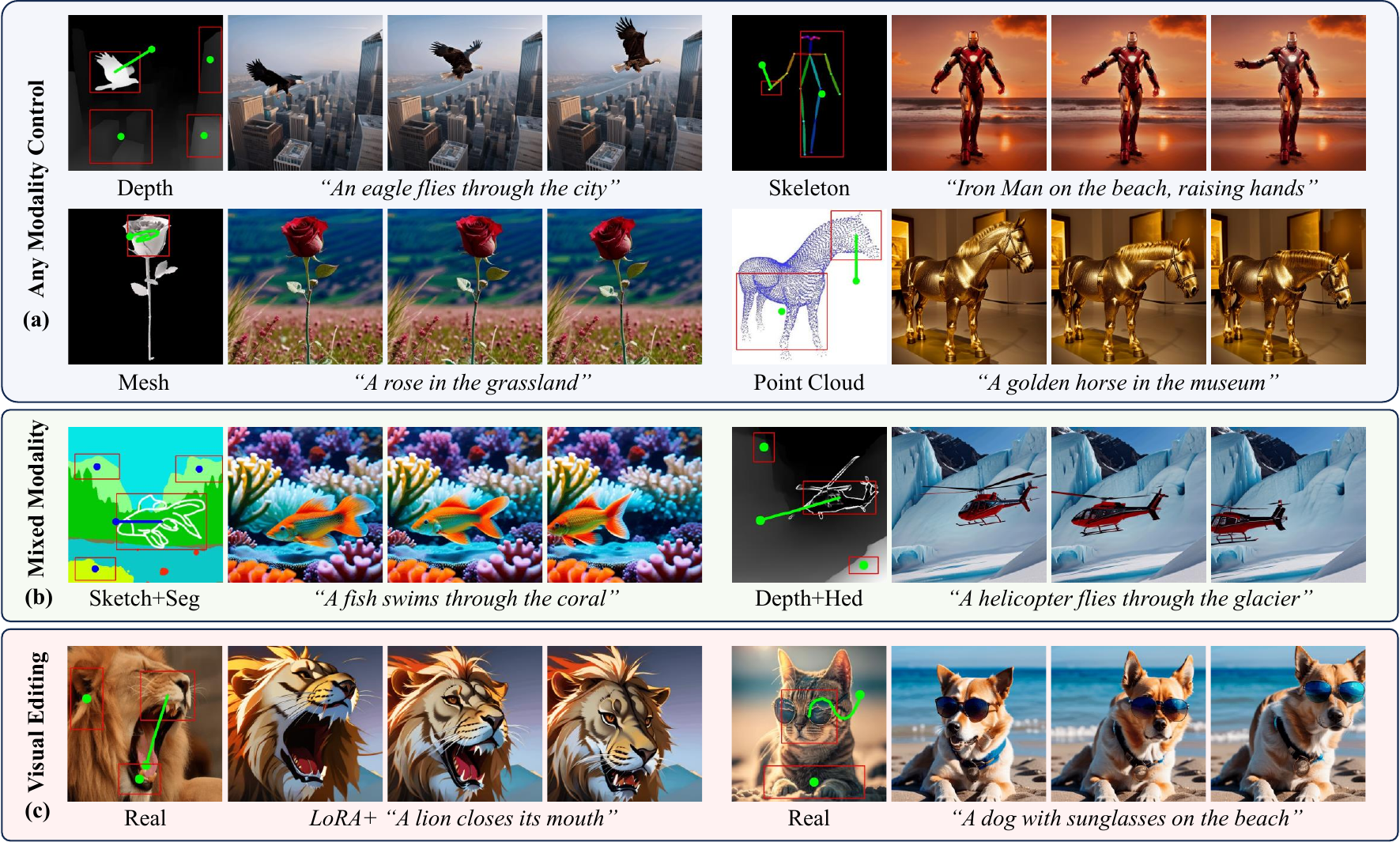}
    \vspace{-6mm}
    \captionof{figure}{The first frame conditional control of our Training-Free architecture \methodname. (a) \methodname supports diverse types of conditional inputs, including those that are difficult to obtain construct pairs for training, such as mesh and point cloud data. The trajectories serve as input for motion control in subsequent frames. (b) \methodname can accept inputs with mixed conditional types, further increasing the flexibility of the input. (c) By using LoRA~\cite{hu2021lora} or different text prompts, \methodname can achieve the editing effect of the original image.}
    \label{fig:teaser}
    \vspace{3mm}
\end{center}%
}]

\renewcommand{\thefootnote}{\fnsymbol{footnote}}
\footnotetext[0]{${\textrm{\Letter}}$ Henghui Ding (henghui.ding@gmail.com) is the corresponding author with
the Institute of Big Data, College of Computer Science and Artificial Intelligence, Fudan University, Shanghai, China.}

\begin{abstract}

Recent advancements in video generation, particularly in diffusion models, have driven notable progress in text-to-video (T2V) and image-to-video (I2V) synthesis. However, challenges remain in effectively integrating dynamic motion signals and flexible spatial constraints. Existing T2V methods typically rely on text prompts, which inherently lack precise control over the spatial layout of generated content. In contrast, I2V methods are limited by their dependence on real images, which restricts the editability of the synthesized content. Although some methods incorporate ControlNet to introduce image-based conditioning, they often lack explicit motion control and require computationally expensive training. To address these limitations, we propose AnyI2V, a training-free framework that animates any conditional images with user-defined motion trajectories. AnyI2V supports a broader range of modalities as the conditional image, including data types such as meshes and point clouds that are not supported by ControlNet, enabling more flexible and versatile video generation. Additionally, it supports mixed conditional inputs and enables style transfer and editing via LoRA and text prompts. Extensive experiments demonstrate that the proposed AnyI2V achieves superior performance and provides a new perspective in spatial- and motion-controlled video generation.
\vspace{-0.4cm}
\end{abstract}

\section{Introduction}
\label{sec:intro}

Diffusion-based video generation has achieved significant advancements recently, with existing methods broadly categorized into text-to-video (T2V) \cite{singer2022make,guo2023animatediff,khachatryan2023text2video} and image-to-video (I2V) \cite{chen2023videocrafter1,wang2024lavie,xing2025dynamicrafter,blattmann2023stable}. Although considerable efforts have been made toward incorporating additional motion signals, e.g., object or camera motion~\cite{wang2024motionctrl}, the integration of motion information with general structural information remains insufficiently explored.

Some T2V-based motion control methods enable object-specific motion control using bounding boxes or point trajectories~\cite{wang2024motionctrl, ma2024trailblazer, yang2024direct, wu2024motionbooth}. While these methods can generate videos following given trajectories, they lack explicit spatial layout control of the generated content, limiting their capability to perform fine-grained adjustments, such as precisely articulating human limbs. In contrast, I2V-based methods~\cite{wang2024boximator,yin2023dragnuwa,shi2024motion} address this limitation by conditioning on the first-frame image, thus explicitly specifying the initial content and achieving finer-grained control. However, the dependence of these methods on real RGB images as first-frame inputs restricts their content editability. To alleviate the challenge, several video diffusion methods~\cite{hu2023videocontrolnet,zhang2023controlvideo} incorporate ControlNet~\cite{zhang2023adding}, injecting frame-wise or sparse conditions~\cite{guo2025sparsectrl} to achieve fine-grained layout control. Nonetheless, these approaches do not support user-defined motion control.~Moreover, both ControlNet and motion control modules require computationally expensive training, and retraining is necessary whenever the base model changes, further limiting their flexibility.

To address these limitations, we propose \methodname, a training-free method that accepts any modality image, \eg, \textit{mesh}, \textit{point cloud}, \textit{edge}, \textit{depth}, \textit{skeleton}, \etc, to serve as a reference frame for initial content layout control while enabling user-defined trajectories for motion control in subsequent frames, as shown in \cref{fig:teaser}.~\methodname is a unified framework that processes the input within one model without additional modules. Beyond single-modality guidance, \methodname also supports mixed-modality inputs, enhancing flexibility and control. For example, depth maps effectively represent background structures, while sketches precisely define foreground details, leveraging the complementary strengths of different modalities. Additionally, by employing LoRA~\cite{hu2021lora} or different text prompts,  \methodname enables editing the visual content of the input image.

Specifically, the proposed \methodname consists of three key methods: structure-preserved feature injection, across-frames alignment, and semantic mask generation. First, to achieve first-frame guidance in a training-free manner, we identify the essential features that should be injected as guidance. Then, to mitigate bias in specific features, we propose a simple yet effective operation that suppresses unwanted appearance information while preserving the structural integrity of the input image. Based on these essential features, we further analyze their temporal characteristics by visualizing the principal components through PCA dimensionality reduction~\cite{wold1987principal}. Our analysis reveals distinct attributes across these features, highlighting that the query in spatial self-attention maintains strong temporal consistency and an entity-aware semantic representation across frames. Motivated by this key insight, we propose aligning the query from spatial self-attention across different frames to ensure temporally coherent video synthesis. Furthermore, to precisely control irregularly shaped objects during this alignment, it is common practice to apply a mask that excludes unrelated content from optimization. However, since the target object may undergo deformation across frames, the use of such a static mask inevitably restricts the flexibility of object motion. To overcome this limitation, we leverage semantic information embedded within the features and cluster it into a dynamic mask, which can adaptively accommodate object deformation while maintaining precise spatial control.

In summary, our main contributions are as follows:
\begin{itemize} 

\item We introduce \textbf{\methodname}, which integrates the first-frame spatial condition with user-defined trajectory to control the content layout and motion, respectively. Moreover, our training-free framework eliminates the training burden and simplifies adaptation across different backbones.

\item \methodname offers exceptional flexibility, allowing to accept a wide range of conditional images as input for the first frame. Furthermore, \methodname supports mixed-condition inputs and by incorporating LoRA or different text prompts, enables effective visual editing, producing highly diverse and visually appealing results.

\item By rethinking feature injection and employing zero-shot trajectory control with semantic masks, \methodname demonstrates superior performance across diverse scenarios, as validated through extensive experiments, highlighting the effectiveness of our approach.

\end{itemize}

\begin{figure*}[htbp]
  \centering
  \includegraphics[width=0.99\linewidth]{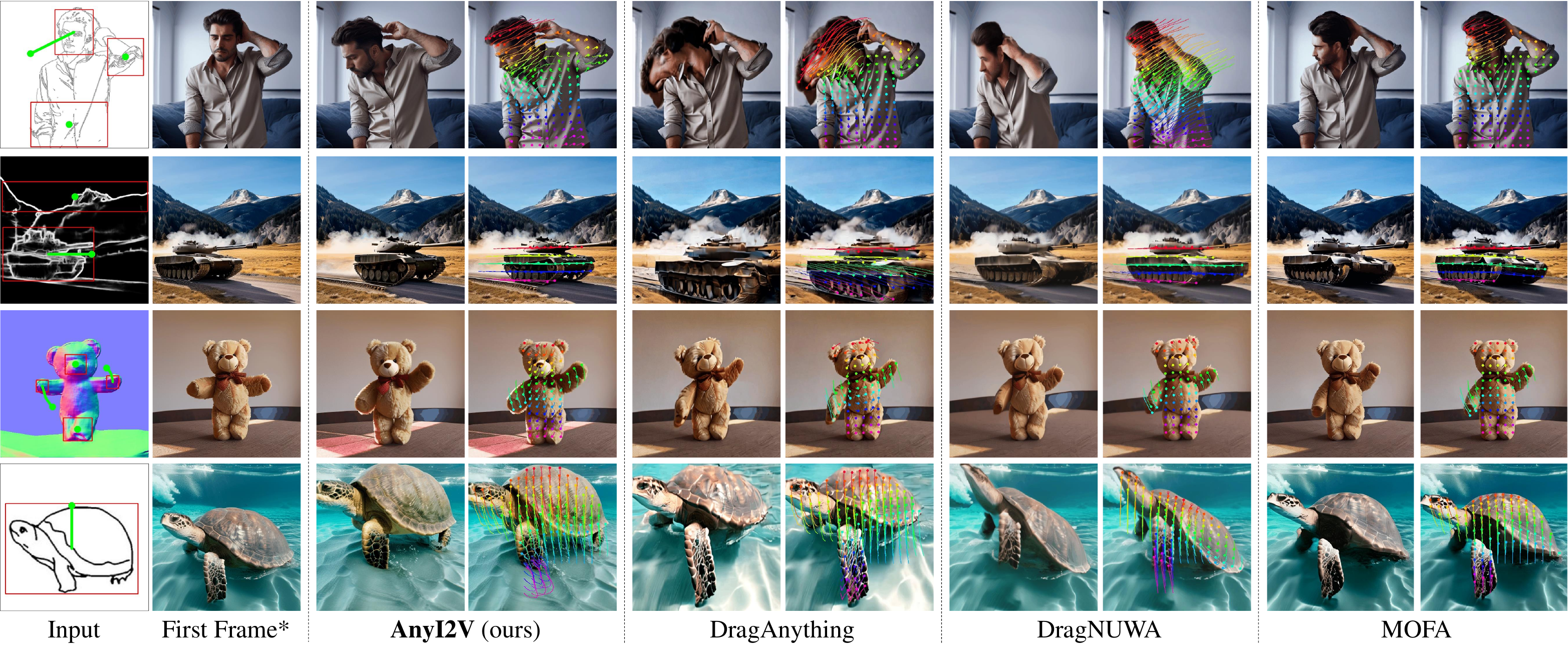}
  \vspace{-3.6mm}
  \caption{Comparison between \methodname and previous methods, DragAnything~\cite{wu2024draganything}, DragNUWA~\cite{yin2023dragnuwa}, and MOFA~\cite{niu2024mofa}. ‘First Frame*’ indicates that the condition images for previous methods are generated using \methodname to ensure a more consistent and fair comparison.}
  \label{fig: comparison}
  \vspace{-0.3cm}
\end{figure*}

\section{Related Work}
\label{sec:related}
\textbf{Diffusion-Based Video Generation.} Generation quality in the field of video generation has achieved great progress in recent years, especially in the domain of diffusion models~\cite{singer2022make, ho2022video,guo2023animatediff,chen2023videocrafter1,wang2024lavie,blattmann2023stable,xing2025dynamicrafter,shuai2024survey}. The earliest video diffusion models focus on text-to-video (T2V) generation, and many of these methods adopt a pre-trained text-to-image (T2I) diffusion model~\cite{rombach2022high} to add motion to the images~\cite{singer2022make,guo2023animatediff,khachatryan2023text2video} where the majority of methods train extra zero-initialized temporal modules while keeping the rest of the modules to be frozen. The later works~\cite{chen2023videocrafter1,wang2024lavie,xing2025dynamicrafter,blattmann2023stable} explore to inject an image as the first frame to provide the spatial layout of the early stage of the generated videos which works as image-to-video (I2V) generation. These methods are mainly built upon the pre-trained T2V model which fine-tune the base model to achieve the ability to accept an image as the condition. 
Although they achieve high-fidelity video generation, the motion in the generated videos is sometimes limited in range or fails to match user expectations. 

\noindent\textbf{Spatial Condition in Diffusion Models.} To control the spatial layout of generation results, several methods~\cite{zhang2023adding,hertz2022prompt,mo2024freecontrol,tumanyan2023plug} have been proposed. Among these, ControlNet~\cite{zhang2023adding} stands out as a groundbreaking approach that accepts input images from a variety of modalities to generate precisely controlled outputs. Building upon this foundation, following methods~\cite{hu2023videocontrolnet,liao2023lovecon,zhang2023controlvideo,he2024cameractrl,wang2024motionctrl,guo2025sparsectrl,mo2024freecontrol} have adapted ControlNet for video generation tasks. However, these methods typically require training the ControlNet with highly aligned datasets. Obtaining aligned inputs and corresponding real images is challenging for some modalities, \eg, meshes or point clouds. Additionally, each modality-specific input demands an independent ControlNet, and the originally pre-trained weights may not be compatible with a new base model. To address these limitations, the recently proposed FreeControl~\cite{mo2024freecontrol} introduces a training-free approach to solve these issues within the text-to-image (T2I) domain. Furthermore, several works~\cite{hu2023cocktail,qin2023unicontrol,lee2024compose,kim2023diffblender,sun2025anycontrol} have explored how to process mixed-modality inputs in the domain of T2I generation, thereby enhancing the diversity and richness of generated content.

\noindent\textbf{Motion Control in Diffusion Models.} Controlling motion in generated videos has been an active area of research~\cite{wang2024motionctrl,he2024cameractrl,cai2021unified,wang2024videocomposer,yin2023dragnuwa,wang2024boximator,zhou2024trackgo,qiu2024freetraj,zhang2024tora,shi2024motion,pan2023drag,shi2024dragdiffusion,yang2024direct,shuai2025free}. Methods such as DragGAN~\cite{pan2023drag} and DragDiffusion~\cite{shi2024dragdiffusion} introduced pioneering approaches to control the destination of a target point in the original image by optimizing the input latent. These methods are implemented in T2I base model and lack smooth temporal transitions. Other methods, such as MotionCtrl~\cite{wang2024motionctrl}, extend motion control to T2V generation, allowing for independent control of camera motion and object motion. Additionally, drag-based methods in I2V generation, such as DragNUWA~\cite{yin2023dragnuwa} and DragAnything~\cite{wu2024draganything}, use an initial image to control the expected motion of objects. Despite their effectiveness, these drag-based T2V and I2V methods often require extensive training data, which can be difficult to obtain, and are implemented in a black-box manner, lacking interpretability. To address these limitations, some training-free methods~\cite{xiao2024video,namekata2024sg} have emerged. However, these approaches only accept input from real image modality and lack an editable space to control the appearance of the video using text prompts. In contrast, this paper proposes a novel method that accepts input from various image modalities and is implemented in a training-free manner, enabling users to edit videos flexibly without the need for fine-tuning the base model.
\begin{figure*}[htbp]
    \centering
    \includegraphics[width=0.99\linewidth]{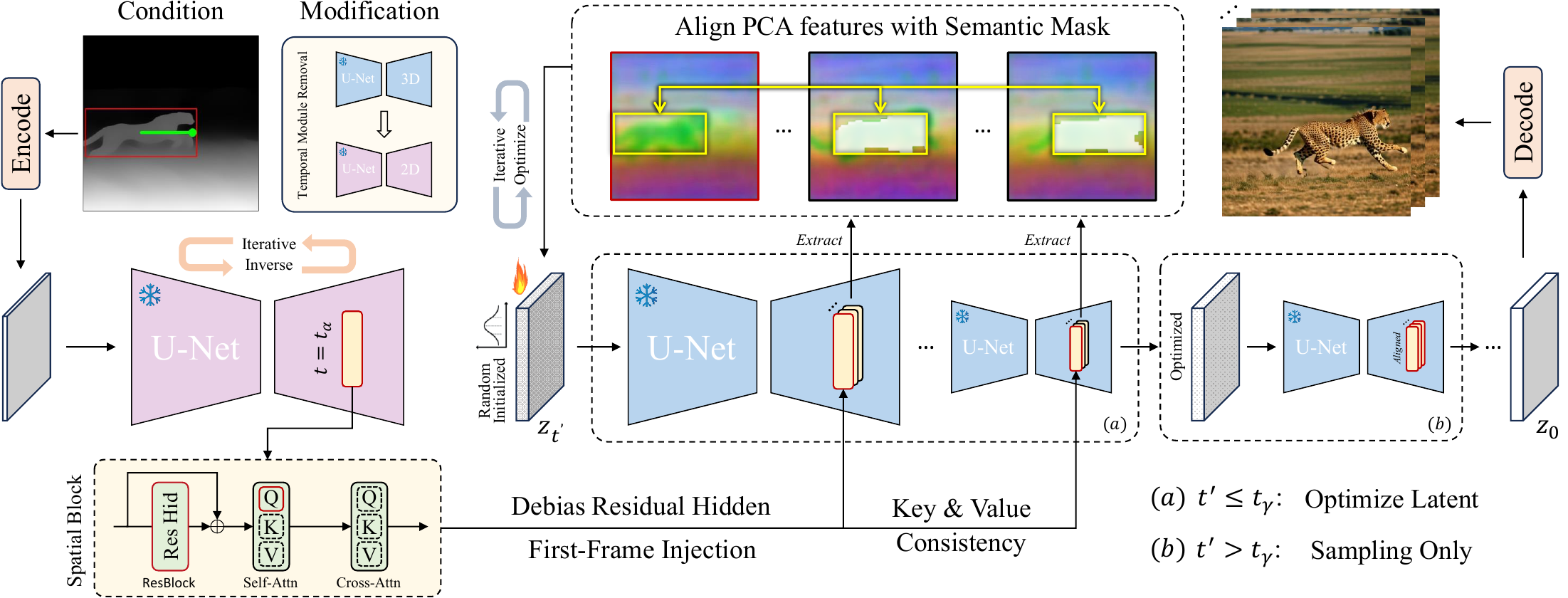}
    \vspace{-0.2cm}
    \caption{\textbf{Overview of Our Pipeline}:
Our pipeline begins by performing DDIM inversion on the conditional image. To do this, we remove the temporal module (i.e., temporal self-attention) from the 3D UNet and then extract features from its spatial blocks at timestep $ t_{\alpha} $. Next, we optimize the latent representation by substituting the features from the first frame back into the U-Net. This optimization is constrained to a specific region by an auto-generated semantic mask (detailed in \cref{fig: mask generation}) and is only performed for timesteps $ t' \leq t_{\gamma} $.}
    \label{fig: pipeline}
    \vspace{-0.08cm}
\end{figure*}

\section{Method}
\label{sec:Method}

\begin{figure}[htbp]
    \setlength{\abovecaptionskip}{3pt}
    \centering
    \includegraphics[width=0.99\linewidth]{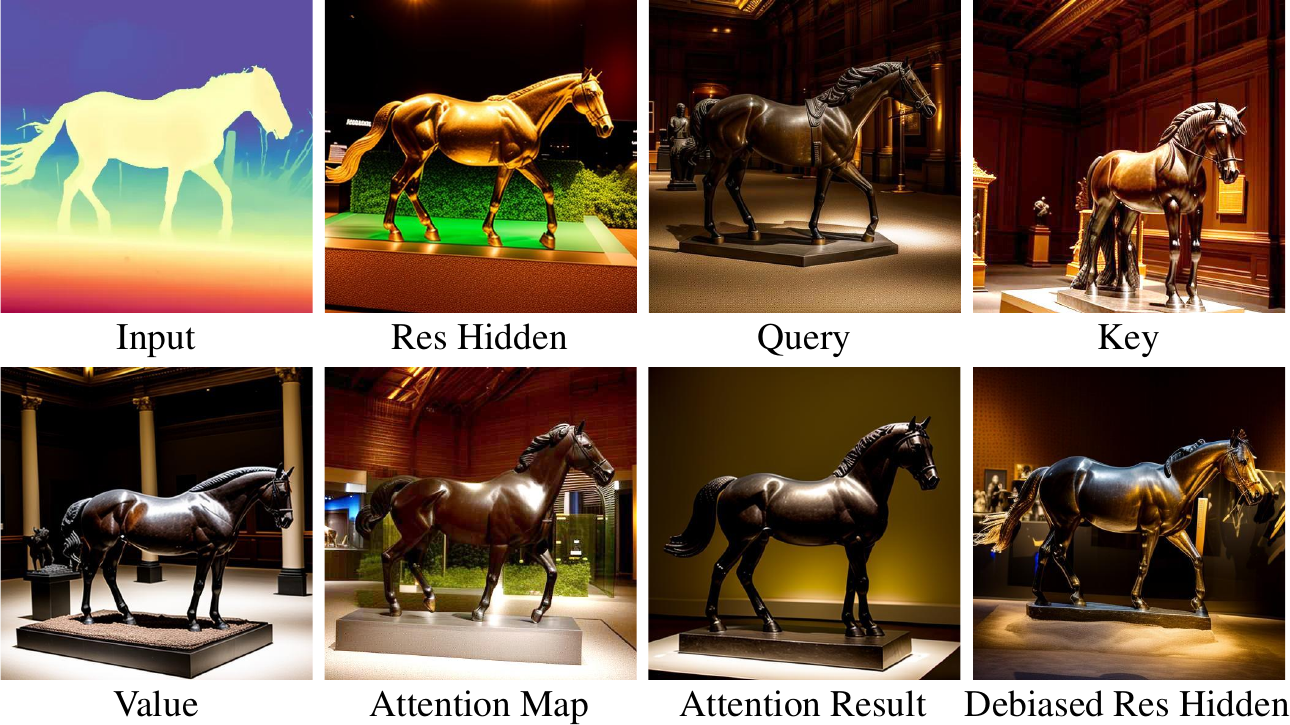}
    \caption{The study examines the influence of injecting different features. Each feature is injected independently. The input text prompt is ``\textit{A sculpture of a horse in the musuem}."}
    \label{fig: feature injection}
    \vspace{-0.15cm}
\end{figure}

\subsection{Preliminary}

\textbf{Architecture.} The proposed \methodname is adapted from a video diffusion model based on a 3D U-Net~\cite{guo2023animatediff}, with spatial and temporal components for frame content and coherence. To leverage pretrained image diffusion models, most T2V methods extend 2D models into 3D by adding temporal modules~\cite{blattmann2023align, guo2023animatediff, he2022latent, wang2023modelscope, chen2023videocrafter1}, and are trained on both images and videos to maintain single-frame generation ability.

\noindent\textbf{Latent Optimization.} Latent optimization involves iteratively updating the latent variable $ z $ by computing gradients of the guidance loss function $ \mathcal{L}_{\text{guide}} $ with respect to $ z $. Through this iterative process, the latent variable $ z $ is progressively refined to better align with the conditioning signals, ensuring that generated samples achieve higher fidelity and accurately reflect the desired conditions.

\begin{figure*}[htbp]
    \setlength{\abovecaptionskip}{3pt}
    \centering
    \includegraphics[width=0.99\linewidth]{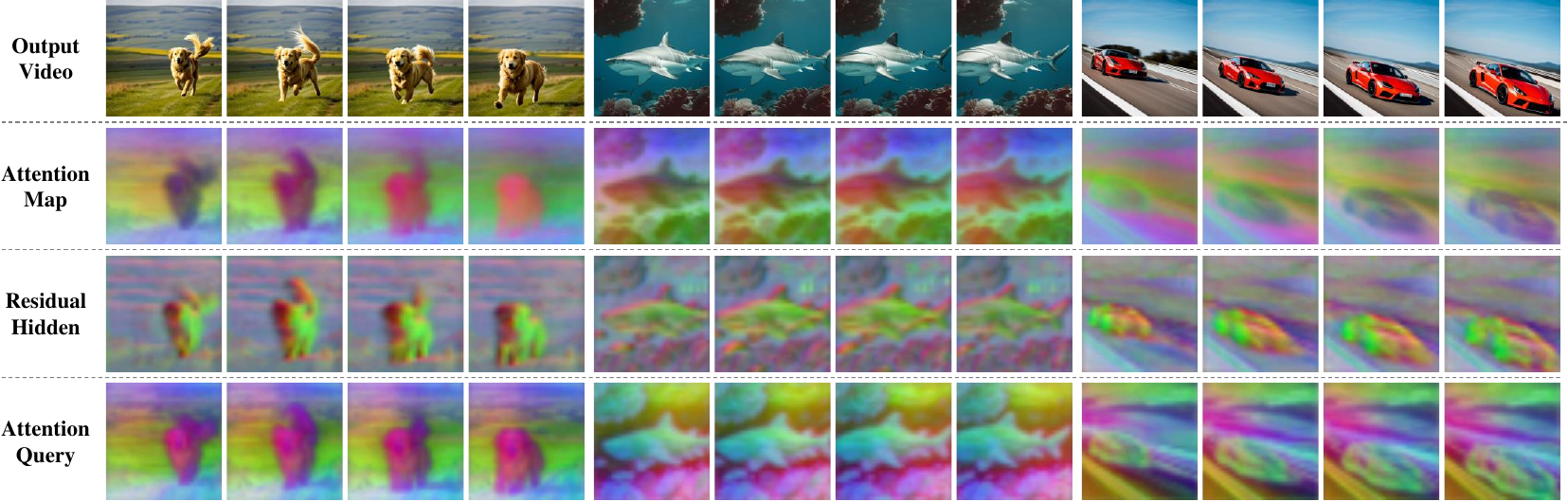}
    \caption{This figure compares different PCA-reduced features in terms of their temporal consistency and entity representation. The results reveal several key observations: (1) Features of attention map exhibit low temporal consistency. (2) Features of residual hidden state fail to treat the object as a coherent entity. (3) The attention query provides both high temporal consistency and strong entity representation.}
    \label{fig: temporal consistency}
    \vspace{-0.2cm}
\end{figure*}

\subsection{Rethinking Feature Injection}

Given an image, it has been demonstrated in PnP~\cite{tumanyan2023plug} that diffusion models present the capability to capture structural information. However, PnP encounters difficulties when processing images from different modalities. Building on this, we conduct experiments to evaluate structural and appearance control by replacing the feature.

As indicated in previous studies~\cite{tumanyan2023plug,shi2024dragdiffusion}, diffusion models typically determine the overall layout in the early denoising stages, while finer details are established in later stages. This observation suggests there is a specific step at which the diffusion model’s features strike an optimal balance between structural information and texture. Accordingly, we extract the feature from a particular DDIM inversion~\cite{song2020denoising} step $t_{\alpha}$ for injection.

Next, we examine how various features at time step $t_{\alpha}$ contribute to the generation result. Instead of using the noise from the end of the DDIM inversion, which retains extensive appearance information, we start with pure random noise and inject only one type of feature at a time. \cref{fig: feature injection} intuitively demonstrates the primary influence of features from the ResBlock and the spatial self-attention layer. Notably, each of the residual hidden, query, and self-attention map independently yields satisfactory structure control, with the residual hidden being especially effective. However, the residual hidden also encapsulates most of the appearance information from the source image leading to unsatisfactory visual fidelity. Hence, to ensure the generated image adheres more closely to the textual guidance, we debias the appearance in the residual hidden, resulting in a more contextually accurate outcome.

Adaptive Instance Normalization (AdaIN)~\cite{huang2017arbitrary} is a widely used technique for maintaining source structure and transferring target style. However, original AdaIN concentrates on operating global features, which leads to inferior result in local quality. To address this issue, we propose to patchify the injected residual hidden feature $ h_{i} $ and the source residual hidden $ h_{s} $ from the backbone into non-overlapping patches $ h'_{i} $ and $ h'_{s} $.  The patchified features are then manipulated using AdaIN to obtain the target hidden representation $ h_{t} $. Formally, the operation is expressed as:
\begin{equation}
h' = \text{Patchify}(h, p), \quad  h_{t} = \text{AdaIN}(h'_{i}, h'_{s}),
\end{equation}
where
    $ h \in \mathbb{R}^{B \times C \times H \times W} $ is the input feature,
    $ h' \in \mathbb{R}^{B \times \left(\frac{H \times W}{p^2}\right) \times C \times p \times p} $ is the patchified feature and $p$ denotes the patch size. The AdaIN is expressed as:
\begin{equation}
\text{AdaIN}(h'_{i}, h'_{s}) = \sigma(h'_{s}) \left( \frac{h'_{i} - \mu(h'_{i})}{\sigma(h'_{i})} \right) + \mu(h'_{s}),
\end{equation}
where $ \sigma $ and $ \mu $ denote functions that compute the standard deviation and mean, respectively, across the spatial dimensions. After the AdaIN operation, the target feature $ h_t $ is reshaped back to its original dimensions to match the input feature map $ h $. As shown in \cref{fig: feature injection}, the debiased residual hidden exhibits well-preserved structure and natural appearance. In this way, the feature injection operation can effectively process images from different modalities while preventing appearance leakage, preserving both structure and appearance fidelity.

To extend feature injection for controlling the first frame in the video diffusion model, we first perform DDIM inversion \cite{song2020denoising} on a single frame conditional image to extract its features. Based on our observation in \cref{fig: feature injection}, we then replace these features by injecting both the debiased residual hidden states and the query. The rationale for using the query instead of self-attention map is discussed in \cref{sec:trajectory_control}. Next, to ensure content consistency across frames in spatial self-attention, we enforce temporal coherence by setting the keys and values of subsequent frames to match those of the first frame, \ie, $K_{2:f}=K_{1}$ and $V_{2:f}=V_{1}$. These strategies not only reduce the computational cost of acquiring target features but also maintains structural control and a natural appearance for the first frame.

\subsection{Zero-Shot Trajectory Control} \label{sec:trajectory_control}
The previous section discussed features without considering the temporal axis. In this section, we select the features with well-structured control ability as discussed in the previous section and use PCA dimensionality reduction to further analyze their characteristics in the temporal dimension. These features are visualized using the first three principal components of the PCA-transformed features. 
As shown in Fig.~\ref{fig: temporal consistency}, we compare the dimensionally reduced features of the self-attention map, residual hidden state, and attention query, focusing on the moving object to assess its temporal consistency and entity representation.

Our findings reveal that moving object in attention map exhibits lower temporal consistency, whereas residual hidden state and attention query demonstrate a strong correlation along the temporal axis of the moving object. Additionally, residual hidden features capture finer-grained details, which do not treat the object as a coherent entity, whereas query features encode the higher-level semantics, treating the object as a whole. This observation leads to a key insight: \textit{Aligning temporally-consistent and entity-aware features across frames enables coherent object motion}.
Based on this, we introduce zero-shot trajectory control by aligning the subsequent frames with the injected first frame.

\textbf{Cross-Frame~Alignment.}~Inspired by previous works \cite{pan2023drag,shi2024dragdiffusion}, latents can be optimized to enable dragging effects on a single image. We apply this technique in trajectory control for cross-frame alignment. Specifically, based on our analysis, we choose \textit{query} in spatial self-attention as the aligning target. Previous works conduct latent optimization in a point-dragging manner, restricting the optimization of latent to a small region. However, for more flexible object control, such as moving a specific part of an object or shifting the entire object, we introduce bounding box $\mathcal{B}$, whose size and position in each frame can be defined by the user. Furthermore, since we observed that lower-ranked components exhibit reduced temporal consistency and struggle to define a clear layout, as shown in the supplementary material, we further propose aligning the higher-ranked principal components of the query features extracted via PCA. Finally, the following optimization objective is used to optimize the latents:
\begin{equation}
z_t^* = \arg\min_{z_t} \sum_{i=1}^{n}\sum_{j=2}^{f}\mathcal{L}\Bigl( F_{j}[\mathcal{B}_{j}^{i}],\ \text{SG}\bigl( F_{1}[\mathcal{B}_{1}^{i}] \bigr) \Bigr),
\end{equation}
where \(\mathcal{L}\) denotes the loss function, defined in \cref{eq: loss function}, \(i\) indexes the bounding box groups, \(j\) represents the frame index and the operator \(\text{SG}\) denotes the stop-gradient operation. \(F_{j}[\mathcal{B}_{j}^{i}]\) represents the extracted feature and cropped by the bounding box \(\mathcal{B}_{j}^{i} \in \mathbb{R}^{H_{j}^{i}\times W_{j}^{i}}\). The extracted feature is defined as:
\begin{equation}
F_{j} = \text{PCA}(\text{Query}_{j}, M),
\end{equation}
where \(M\) is the number of principal components of channel dimension. Notably, the feature \(F_{1}\) corresponds to the injected first-frame feature and is independent of \(z_t\).

\subsection{Semantic Mask Generation.} 
The bounding box mentioned above offers flexibility in defining the target region for dragging, yet it does not always enable precise object manipulation. Many objects have irregular shapes, causing unintended regions to be optimized and compromising overall accuracy. Meanwhile, a static mask can further limit the optimized area but also constrains natural deformations, reducing flexibility during dynamic transformations. To overcome these issues, we introduce an adaptive semantic mask generation method that automatically produces a mask based on the semantic information encoded in the features. By doing so, it provides more accurate, context-sensitive, and adaptive control over the target object, preserving structural integrity while allowing for natural movement.

Given the injected first-frame feature $F_{1}$, we aim to generate more precise masks in the all features $F_{1:f}$ corresponding to the semantic content in the $F_{1}$. First, the salient points $P_{i}^{k}$ are selected inside the bounding boxe of feature $F_{1}$ to indicate interesting parts, where $i$ denotes the corresponding group index of bounding box and $k$ represents the point index. For all frames, we compute the similarity between $P_{i}^{k}$ and the features within bounding boxes $\mathcal{B}_{j}^{i}$ using cosine similarity, formulated as follows:
\begin{equation}
    SIM_{i,j}^k = \frac{F_{j}[\mathcal{B}_{j}^{i}] \cdot F_{1}[P_{i}^{k}]}{\|F_{j}[\mathcal{B}_{j}^{i}]\| \|F_{1}[P_{i}^{k}]\|} \in \mathbb{R}^{H_{j}^{i} \times W_{j}^{i}},
\end{equation}
where $F_{1}[P_{i}^{k}] \in \mathbb{R}^{M}$ represents the vector at the coordinate $P_{i}^{k}$ in $F_{1}$. Then we can derive the aggregated similarity map $SIM_{i,j}$ of group $i$ on the frame $j$ as follows:
\begin{equation}
    SIM_{i,j}(h,w) =  \max_{k=1, \dots, K} SIM_{i,j}^k(h, w).
\end{equation}
To obtain a binary mask, we apply K-Means clustering~\cite{macqueen1967some} to the aggregated similarity map:
\begin{equation}
    M_{j}^{i} = \text{KMeans}(SIM_{i,j}, 2) \in \mathbb{R}^{H_{j}^{i} \times W_{j}^{i}},
\end{equation}
where 2 denotes binary clustering.~The foreground is determined by selecting the cluster with the higher-value centered pixel.
Based on the derived mask, we define the loss function as:
\begin{equation}
\label{eq: loss function}
    \mathcal{L}_{j}^{i} = || M_{1}^{i} \odot  M_{j}^{i} \odot (F_{j}[\mathcal{B}_{j}^{i}] - \text{SG}(F_{1}[\mathcal{B}_{1}^{i}])) ||_2^2,
\end{equation}
where $\odot$ denotes pixel-wise multiplication, and $M_{1}^{i} \odot M_{j}^{i}$ indicates the overlapping region of the same instance within the bounding box group $\mathcal{B}^{i}$ between frame 1 and frame $j$.

\begin{figure}[t]
    \centering
    \includegraphics[width=0.95\linewidth]{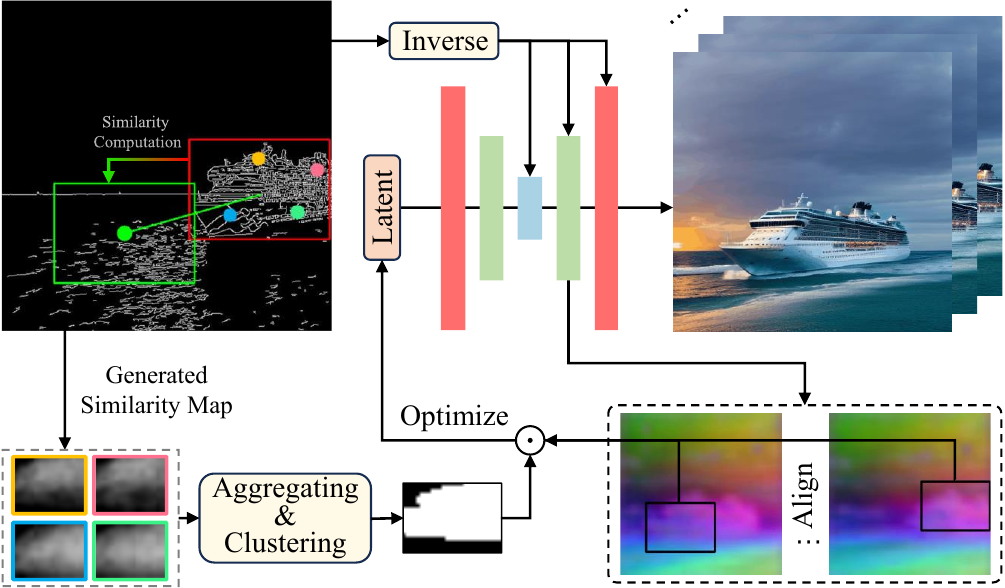}
    \vspace{-1mm}
    \caption{The process of aligning the object across frames by optimizing the latent noise while incorporating semantic masks.}
    \label{fig: mask generation}
    \vspace{-1mm}
\end{figure}

\section{Experiments}

\begin{figure*}[htbp]
    \centering
    \includegraphics[width=0.998\linewidth]{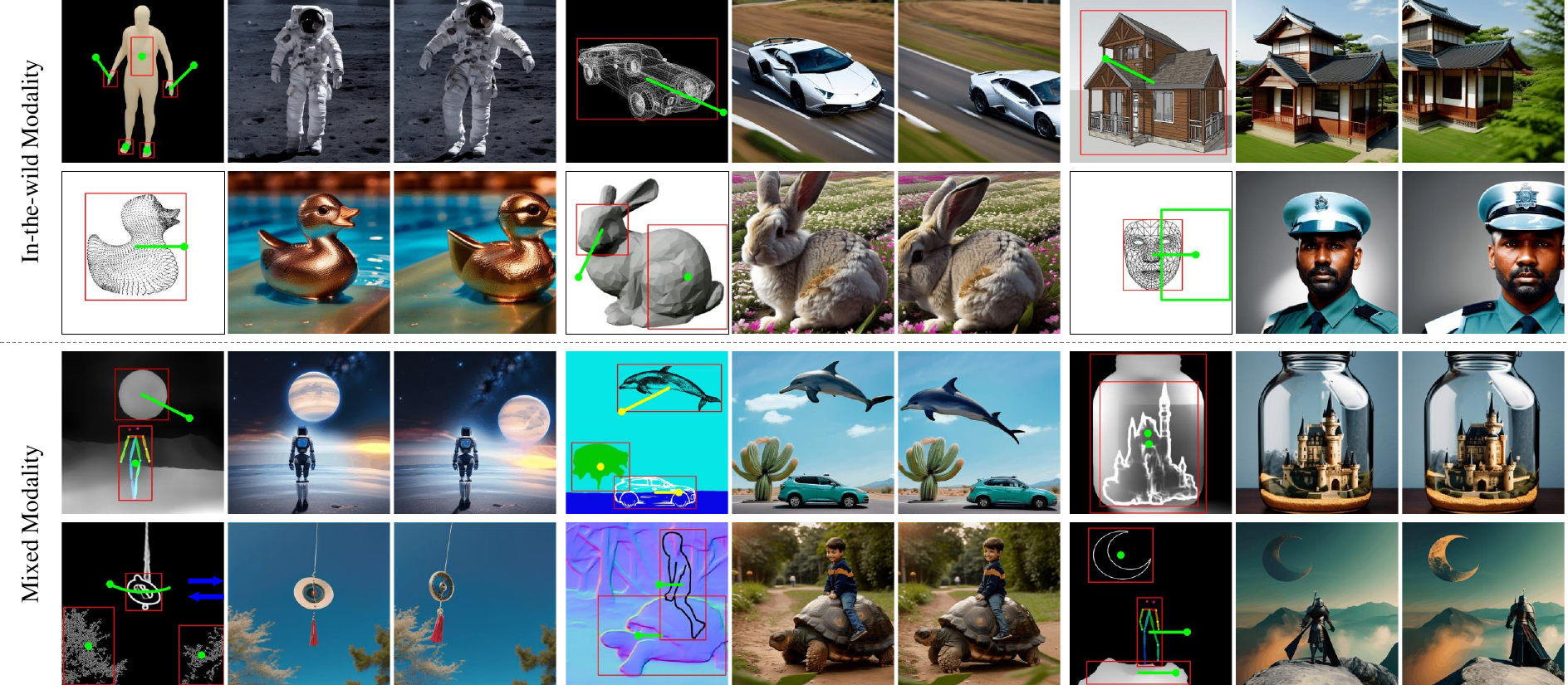}
    \vspace{-6.6mm}
    \caption{This picture demonstrates \methodname's ability to control diverse conditions. \methodname can not only handle modalities that ControlNet does not support but also effectively control mixed modalities, which previously required additional training by other methods.}
    \label{fig: wild&mixed conditions}
    \vspace{-0.3cm}
\end{figure*}

\textbf{Implementation Details.}
We implement our method based on AnimateDiff~\cite{guo2023animatediff} using a single Nvidia A800 GPU. Our overall pipeline can be seen in \cref{fig: pipeline}. The DDIM inversion consists of 1000 steps, with features extracted at $t_{\alpha} = 201$. The decoder has three cascaded spatial blocks, indexed as 0, 1 and 2. We inject residual hidden and query indexed as 0 and 1 from \texttt{up\_blocks.1} and \texttt{up\_blocks.2}, optimizing the latent noise by aligning query 1 from \texttt{up\_blocks.1} and query 0 from \texttt{up\_blocks.2}. The patch window is set to be $p=4$ for debiasing the residual hidden. PCA reduction dimension is set to $M = 64$. We use 25 DDIM sampling steps, optimizing latents every 5 steps for $t' \geq 20$ with a 0.01 learning rate. Inversion stage and generation stage take approximately 8s and 35s in half-precision mode, respectively. 

\subsection{Qualitative Evaluations}
\cref{fig: comparison} presents a comparison between \methodname and previous state-of-the-art methods~\cite{wu2024draganything,yin2023dragnuwa,niu2024mofa,li2024image,wang2024objctrl,qiu2024freetraj,ma2024trailblazer}. From the result images and the visualized trajectories, \methodname demonstrates comparable results. \cref{fig: wild&mixed conditions} further showcases the ability of \methodname to handle in-the-wild and mixed-modality images.~Unlike compared methods that are limited to processing only realistic RGB images, our approach significantly enhances editability and flexibility.

\begin{table}[t]
    \centering
    \footnotesize
    \caption{Comparison with previous state-of-the-art methods.}
    \label{tab:quantitative comparison}
    \vspace{-3mm}
    \setlength\tabcolsep{5.86pt}
    \begin{tabular}{c|c|ccc}
       \specialrule{.1em}{.05em}{.05em} 
       \rowcolor[gray]{.92} Methods & Training & FID $\downarrow$ & FVD $\downarrow$ & ObjMC $\downarrow$ \\
       \hline
       ImageConductor~\cite{li2024image} & \checkmark & \textcolor{gray}{132.23} & \textcolor{gray}{646.50} & \textcolor{gray}{21.14} \\
       DragAnything~\cite{wu2024draganything} & \checkmark & \textcolor{gray}{\underline{95.83}} & \textcolor{gray}{\textbf{556.09}} & \textcolor{gray}{\textbf{13.60}} \\
       DragNUWA~\cite{yin2023dragnuwa} & \checkmark & \textcolor{gray}{105.02} & \textcolor{gray}{\underline{575.78}} & \textcolor{gray}{\underline{15.02}} \\
       MOFA-Video~\cite{niu2024mofa} & \checkmark & \textcolor{gray}{\textbf{95.63}} & \textcolor{gray}{599.48} & \textcolor{gray}{17.72} \\
       \hline
       Baseline~\cite{guo2023animatediff} & -- & 141.95 & 970.26 & 38.26 \\
       FreeTraj~\cite{qiu2024freetraj} & \textbf{Free} & 128.78 & 672.87 & 24.00 \\
       TrailBlazer~\cite{ma2024trailblazer} & \textbf{Free} & 112.37 & 620.80 & 23.71 \\
       ObjCtrl-2.5D~\cite{wang2024objctrl} & \textbf{Free} & \underline{111.82} & \underline{605.96} & \underline{23.12} \\
       \rowcolor{cyan!10}
       \textbf{\methodname} (\textbf{ours}) & \textbf{Free} & \textbf{104.53} & \textbf{569.89} & \textbf{16.39} \\
       \bottomrule
    \end{tabular}
    \vspace{-0.5cm}
\end{table}

\subsection{Quantitative Evaluations}
Following previous work~\cite{wu2024draganything}, we collect data from the web and the VIPSeg dataset~\cite{miao2022large}, annotating video trajectories using Co-Tracker~\cite{karaev2024cotracker} to ensure high-quality motion tracking~\cite{MeViS}. The evaluation metrics include Fréchet Inception Distance (FID), Fréchet Video Distance (FVD), and ObjMC, where ObjMC quantifies the error between the ground truth trajectory and the generated result, providing a fine-grained assessment of motion accuracy.  

For a fair comparison, we randomly convert the first input frame into various structural representations, including canny, HED, depth, normal, and segmentation maps. \methodname directly utilizes these representations, while other methods first process the input frames with ControlNet. As shown in \cref{tab:quantitative comparison}, \methodname significantly outperforms the baseline model and achieves competitive results against state-of-the-art methods. Here, `baseline' refers to the backbone model with SparseCtrl~\cite{guo2025sparsectrl}, which accept the reference image and conduct the experiment in a zero-shot manner (indicated by the `\textcolor{gray}{-}' training attribute).

\begin{table}[t]
    \centering
     \footnotesize
     \caption{Ablation study of the proposed \methodname}
        \label{tab:ablation study}
        \vspace{-3mm}
     \setlength\tabcolsep{10.86pt}
        \begin{tabular}{c|ccc}
           \specialrule{.1em}{.05em}{.05em} 
           \rowcolor[gray]{.92} Options & FID $\downarrow$ & FVD $\downarrow$ & ObjMC $\downarrow$ \\
            \hline
            w/o K\&V consistency & 108.18 & 587.69 & \underline{16.81} \\
            w/o PCA Reduction & 105.95 & 585.04 & 17.14 \\
            w/ Static Mask &  \underline{105.44}  & 598.15 & 16.92 \\
            w/o Semantic Mask & 105.78 & \underline{579.88} & 17.62 \\
            opt. Residual Hidden & 129.40 & 647.52 & 36.23 \\
            \hline
           \rowcolor{cyan!10} \textbf{Full} (\textbf{AnyI2V}) & \textbf{104.53} & \textbf{569.89} & \textbf{16.39} \\
            \bottomrule
        \end{tabular}
        
    \vspace{-0.5cm}
\end{table}

\begin{figure*}[htbp]
    \setlength{\abovecaptionskip}{2pt}
    \centering
    \includegraphics[width=0.98\linewidth]{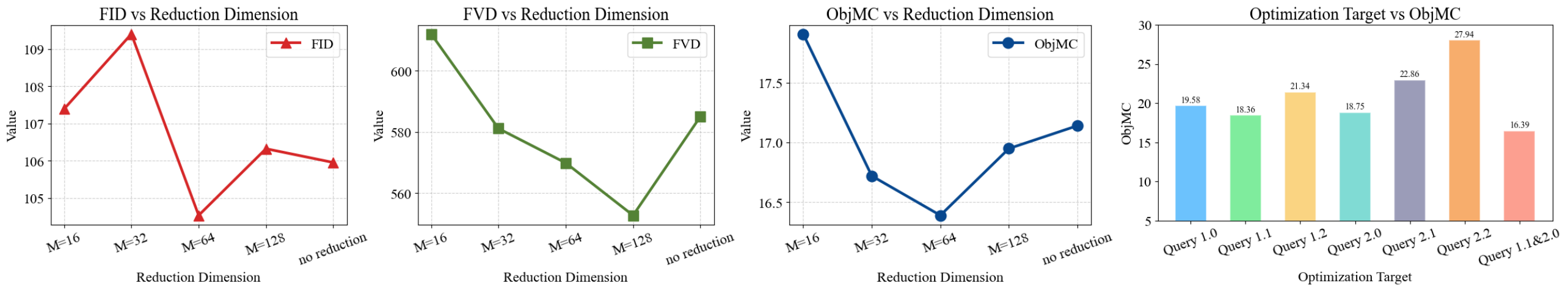}
    \caption{The line chart in the figure demonstrates the ablation study on the impact of different PCA reduction dimensions on FID, FVD, and ObjMC, while the histogram presents the ablation study on optimization targets for ObjMC.}
    \label{fig: line chart}
    \vspace{-0.4cm}
\end{figure*}

\begin{figure}[htbp]
    \centering
    \begin{minipage}{\linewidth}
        \setlength{\abovecaptionskip}{2pt}
        \centering
        \includegraphics[width=0.98\linewidth]{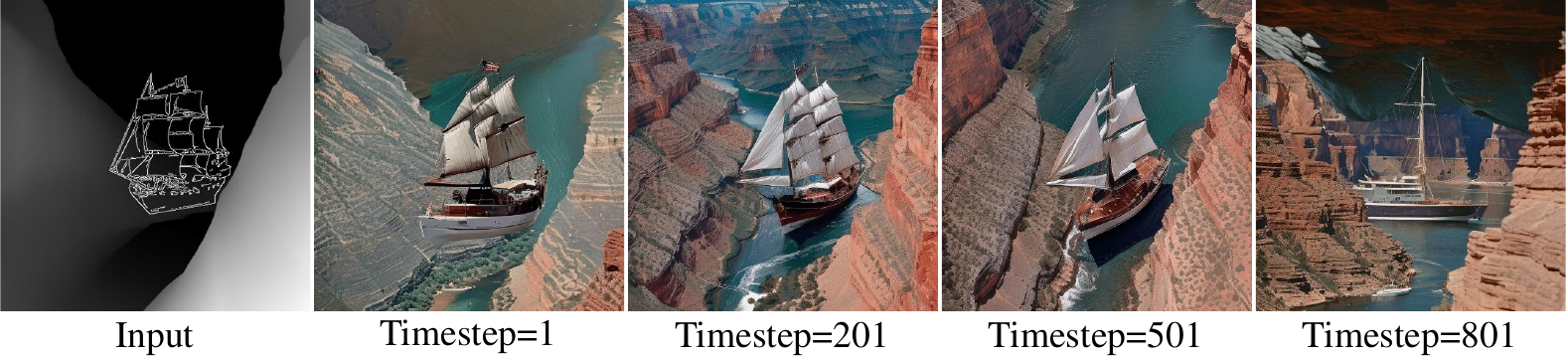}
        \caption{The ablation study examines feature injection across time steps. Visual results are from the first output frames.}
        \label{fig: timestep ablation}
    \end{minipage}
    
    \vspace{2mm}

    \begin{minipage}{\linewidth}
        \setlength{\abovecaptionskip}{2pt}
        \centering
        \includegraphics[width=0.98\linewidth]{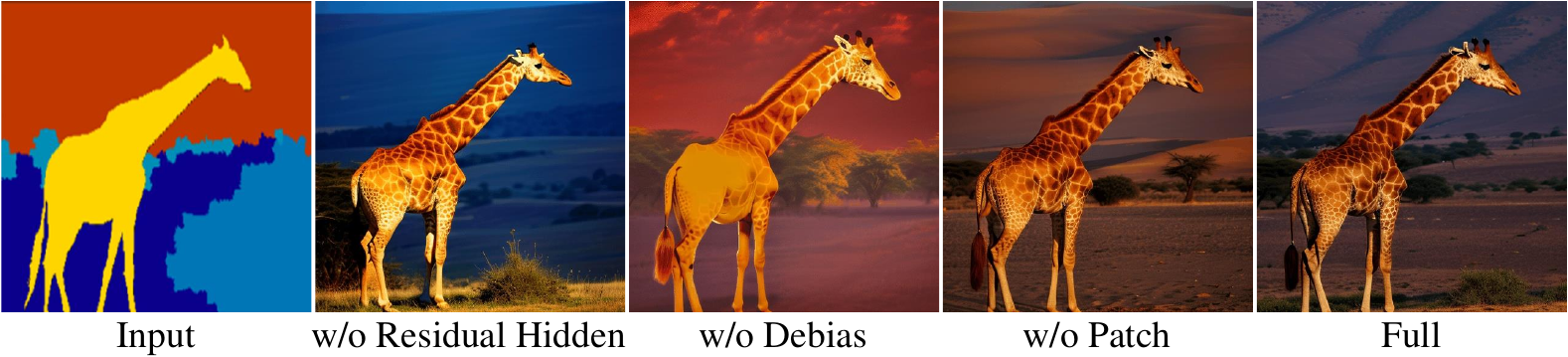}
        \caption{The ablation study examines operations on the residual hidden state. Visual results are from the first output frames.}
        \label{fig: res hidden ablation}
    \end{minipage}
    \vspace{-0.6cm}
\end{figure}

\subsection{Ablation Study}
\cref{tab:ablation study}, \cref{fig: line chart}, \cref{fig: timestep ablation}, and \cref{fig: res hidden ablation} show the ablation study of different design choices. \cref{tab:ablation study} evaluates the impact of removing our proposed components and optimizing the residual hidden states instead of the query in self-attention. The results demonstrate that our proposed configurations enhance the temporal consistency of the generated videos, as reflected in the FVD metric, and improve the precision of target object control according to the ObjMC metric.

The line chart in \cref{fig: line chart} illustrates how PCA reduction dimensions affect FID, FVD, and ObjMC. Both overly small and large dimensions degrade performance: small dimensions leaves insufficient information for effective alignment despite strong temporal coherence, while large ones weaken consistency in lower-ranked components, hindering alignment. Based on this, we select $M=64$ as the optimal PCA dimension for latent alignment.

\cref{fig: timestep ablation} examines the impact of injecting features at different time steps. When the time step is too small, the model tends to overfit low-level textures, leading to unnatural artifacts. When the step is too large, noisy features hinder the model to capture accurate layout information, disrupting the overall structure. An appropriately chosen time step balances low-level feature extraction with layout preservation, resulting in higher-fidelity visuals.

\cref{fig: res hidden ablation} investigates the impact of residual hidden operations. The results indicate that removing the residual hidden leads to poorer control over object details (e.g., giraffe legs). Without debiasing, the model tends to overfit the original input conditions. Moreover, without slicing patches results in incomplete debiasing of appearance features. In contrast, our approach effectively maintains layout control while preventing overfitting to the input image's appearance.

The histogram in \cref{fig: line chart} presents the results of the ablation study on different optimization targets. We index the \texttt{j}-th query in decoder \texttt{up\_blocks.i} as \texttt{Query i.j}. The results show that optimizing a single query, specifically \texttt{Query 1.1} and \texttt{Query 2.0}, yields the leading performance. when optimizing queries from different resolutions, such as \texttt{Query 1.1~\!\&~\!2.0} (our setting), a significant improvement is observed. This enhancement occurs because optimizing queries across multiple resolutions enables the model to capture both semantic and structural information at various levels. For example, conditions like canny and HED often contain hollow regions that may disappear at smaller resolutions. Higher resolutions, however, provide finer texture details. Consequently, Multi-resolution optimization enhances accuracy and visual fidelity.

\begin{figure}[t]
    \setlength{\abovecaptionskip}{2pt}
    \centering
    \includegraphics[width=0.98\linewidth]{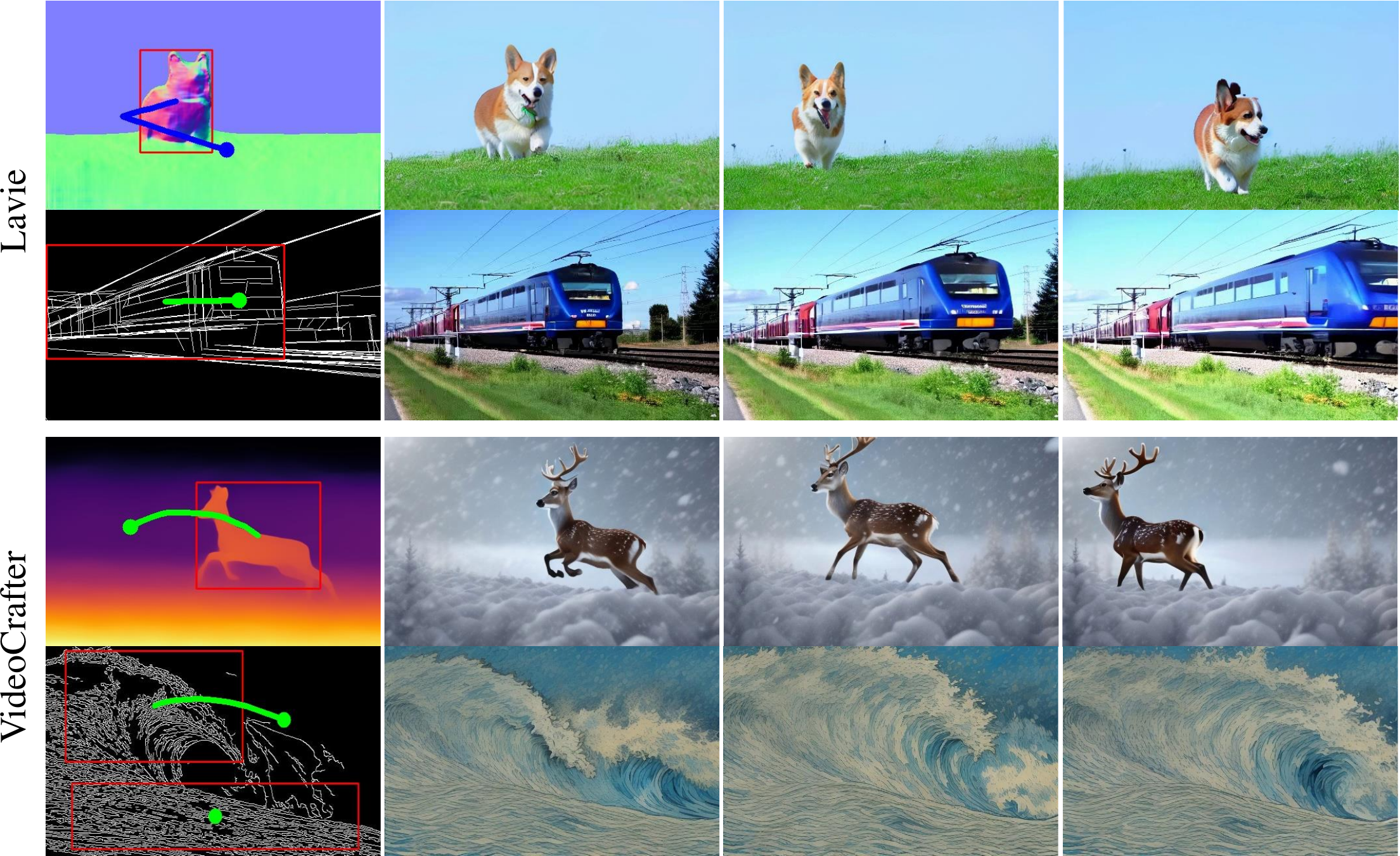}
    \caption{The generalization test results of \methodname on other backbones, including Lavie~\cite{wang2024lavie} and VideoCrafter2~\cite{chen2024videocrafter2}.}
    \label{fig: different backbones}
    \vspace{-0.4cm}
\end{figure}

\subsection{Generalization}

Since \methodname is a training-free method, we further implement it on different T2V backbones, including Lavie~\cite{wang2024lavie} and VideoCrafter2~\cite{chen2024videocrafter2}.~The results shown in \cref{fig: different backbones} highlight its adaptability to various architectures, demonstrating its robustness and strong generalization ability.
\vspace{-0.1cm}

\section{Conclusion}
\vspace{-0.1cm}

We proposed \textbf{\methodname}, a training-free image-to-video (I2V) generation approach adapted from the text-to-video (T2V) backbone that integrates flexible spatial conditions from any modality and motion control using user-defined trajectories. Unlike previous methods, \methodname eliminates the need for extensive training and also simplifies transferring between different backbones, providing convenience for application.

\noindent\textbf{Limitations and Future Work.} Despite its advantages, \methodname has limitations. It struggles with precise control of very large motion ranges and ambiguous occlusions, which can lead to unclear spatial relationships. Additionally, as feature injection occurs only at earlier denoising steps, the first frame lacks the precise control offered by methods like ControlNet. Future work can focus on improving motion consistency, handling complex occlusions, and integrating lightweight fine-tuning for better adaptability.

\footnotesize{\paragraph{Acknowledgement.}~This project was supported by the National Natural Science Foundation of China (NSFC) under Grant No. 62472104. This work was supported by Damo Academy through Damo Academy Innovative Research Program.}
\clearpage
{
    \small
    \bibliographystyle{ieeenat_fullname}
    \bibliography{main}
}

\end{document}